\documentclass[10pt,twocolumn,letterpaper]{article}

\usepackage[pagenumbers]{cvpr} % To force page numbers, e.g. for an arXiv version

\usepackage[dvipsnames]{xcolor}

\definecolor{cvprblue}{rgb}{0.21,0.49,0.74}
\usepackage[pagebackref,breaklinks,colorlinks,citecolor=cvprblue]{hyperref}
\usepackage[most]{tcolorbox}
\usepackage{wrapfig}
\usepackage[margin=1in]{geometry}
\usepackage{color,soul}
\usepackage{subcaption}
\newcommand{\Blue}[1]{{\color{blue}\ul{#1}}}

\def\paperID{21} % *** Enter the Paper ID here
\def\confName{CVPR}
\def\confYear{2024}

\title{On Speculative Decoding for Multimodal Large Language Models}

\author{Mukul Gagrani\thanks{Equal contribution. \\
Correspondence to \{mgagrani,raghgoel,mingul\}@qti.qualcomm.com. \\
Qualcomm AI Research is an initiative of Qualcomm Technologies, Inc. }
\quad
 Raghavv Goel\footnotemark[1] 
 \quad
 Wonseok Jeon 
 \quad
 Junyoung Park 
 \quad
 Mingu Lee 
 \quad
 Christopher Lott \\
 Qualcomm AI Research \\
 }

\begin{document}
\maketitle

\begin{abstract}
Inference with Multimodal Large  Language Models (MLLMs) is slow due to their large-language-model backbone which suffers from memory bandwidth bottleneck and generates tokens auto-regressively. In this paper, we explore the application of speculative decoding to enhance the inference efficiency of MLLMs, specifically the LLaVA 7B model. We show that a language-only model can serve as a good draft model for speculative decoding with LLaVA 7B, bypassing the need for image tokens and their associated processing components from the draft model. Our experiments across three different tasks show that speculative decoding can achieve a memory-bound speedup of up to 2.37$\times$ using a 115M parameter language model that we trained from scratch. Additionally, we introduce a compact LLaVA draft model incorporating an image adapter, which shows marginal performance gains in image captioning while maintaining comparable results in other tasks. 
\end{abstract}

\section{Introduction}

% \begin{itemize}
% \item Multi-modal language models are gaining popularity
% \item Inference is slow due to AR generation and memory bound processing
% \item Speculative decoding has been used to speedup inference for LLMs
% \item No prior work has shown if speculative decoding can be used for multi-modal language models
% \item We show that a language model can serve as a good draft model for multi-modal LLMs without requiring the image adapter and vision encoder pipeline
% \item Evaluation on LLaVA instruction finetuning datatset, ScienceQA, captioning taks on images from coco dataset
% \end{itemize}

% Large vision models (LVMs) are becoming ubiquituos due to their practical and potential use-cases in medicine, search, undertstanding, etc. (cite paper for each). Given the popularity of Large Language Models (LLMs), a natural extension is to use images as well. Images have the ability to describe different concepts/ideas for which a lot of language tokens might be needed. Thus, having the capability of using both image and language tokens will improve the model's capability in interacting with human user by better responses. 
% The large multimodal models use LLMs are the base language model and attach a separate vision tower for encoding the image. The vision tower mostly based on Vision transformer (ViT) (cite) and an image projection/adapter layer to project the image representations to the hidden-dimension of the language model. The LMMs, thus, incorporates both the benefits of an already trained language model and also its shortcomings: auto-regressive generation. 

Large Language Models (LLMs) have become ubiquitous across various domains due to their impressive performance. However, LLMs only take text queries as input but real-world data comes in the form of multiple modalities including visual data. Multi-modal Large Language Models (MLLMs) \cite{awadalla2023openflamingo, liu2024visual, tsimpoukelli2021multimodal, zhu2023minigpt} provides the LLMs with image understanding abilities, and the fusion of visual and textual tokens enhances the model’s interaction with users, leading to more informed responses. MLLMs comprise of an image encoder to process the image information and an adapter which transforms the image encodings to the language model embedding space. In addition, MLLMs have a language-model backbone in the form of a LLM and thus inherit the auto-regressive generation and memory-bandwidth bottleneck which lead to slow inference \cite{shazeer2019fast}. 

Speculative decoding \cite{leviathan2023fast, chen2023accelerating, sun2023spectr, miao2023specinfer, jeon2024recursive} has been proposed as a solution to accelerate the LLM inference without loss in accuracy, where a smaller draft model predicts multiple future tokens which are verified in a single call of the LLM. Given that MLLMs have a LLM backbone, speculative decoding can be used to make inference with MLLMs more efficient. Many recent works have studied the application of speculative decoding and its variants \cite{kim2023big,fu2023lookahead,medusa,santilli2023accelerating,sun2023spectr,jeon2024recursive} for LLMs, but no such work exists in the context of MLLMs to the best of our knowledge.

In this paper, we apply speculative decoding to LLaVA 7B model (with LLaMA 7B model as language-model backbone) to make inference more efficient, block diagram shown in Figure \ref{fig:SPD_block_diagram}. Due to the lack of publicly available models of LLaVA and LLaMA families smaller than 7B parameters, we train a language model of  size 115M from scratch for speculative decoding. We show that language-only model which does not consider the image tokens (and hence does not require the image encoder and adapter) can serve as a good draft model for LLaVA 7B. We conduct experiments on three different tasks including image QA on LLaVA Instruct 150K dataset \cite{liu2024visual}, image captioning on Coco dataset \cite{lin2014microsoft} and ScienceQA dataset \cite{lu2022learn}, using draft model candidates which have gone through different stages of training and fine-tuning. Our results show that we can achieve memory-bound speedup of upto 2.37$\times$ using only a language model as draft model. We also create a small LLaVA draft model which consists of an image adapter along with our trained language model and show that it improves the performance slightly on COCO captioning task and ScienceQA task while performing similar to language-model-only draft models on the other tasks.

% As humans, we understand we our understanding depends on what we see and what we communicate, thus we use images and language to understand the world and also make decisions. 

% \section{Method}

% - Speculative decoding preliminaries 
% - Draft model training
% - Add a schematic showing the inference pipeline with spec decoding

\section{Background}

\subsection{Speculative Decoding}
SPeculative Decoding (SPD)~\cite{chen2023accelerating,leviathan2023fast} involves a smaller draft model generating multiple tokens which are verified in parallel by the target LLM. Given an input context $X_{1:n}:=[X_{1}, \dots, X_{n}]$, the draft model generates a sequence of tokens $\hat{X}_{n+1:n+L}$ in an auto-regressive fashion, $\hat{X}_{n+j} \sim p(\cdot | X_{1:n}, \hat{X}_{n+1:n+j-1})$. The draft tokens are then verified via a single call of the target LLM ($q$) using rejection sampling criteria that guarantees the same output token distribution as that of the target LLM. Specifically, token $\hat{X}_{n+j}$ is accepted with probability 
\begin{align*}
    \min\left\{1, \frac{q(\hat{X}_{j}|X_{1:n}, \hat{X}_{n+1:n+j-1})}{p(\hat{X}_{j}|X_{1:n}, \hat{X}_{n+1:n+j-1})}\right\}.   
\end{align*}
If a draft token $\hat{X}_{n+j}$ is rejected, then a new token is sampled from the residual distribution defined as $p_{res}(x)=\max(0, q(x) - p(x) )$. 

\subsection{Multimodal Large Language Models}
An image-based Multimodal Large Language Model (MLLM) consists of \textit{1) a vision encoder} to encode the input image, \textit{2) an adapter} to convert the image encodings to language model embeddings, and \textit{3) a language-model backbone}. We describe the framework of the LLaVA model in more detail as follows; given an input image $I$ and the text query $Q$, the image $I$ is converted into a sequence $H_1, H_2, \ldots, H_m$ of $m$ image encodings, and the text query is converted to a sequence of token embeddings $X_1, X_2, \ldots X_n$. The image encodings are further transformed via an adapter $g_\theta$ (a small multi-layer perceptron) to get image embeddings, $V_i = g_\theta(H_i)$. This is done to convert the encodings $H_i$ to the language model embedding space. Tokens are then generated by the language model conditioning on the image embeddings and the token embeddings as follows: 
\begin{equation}
    X_{n+1} \sim q(\cdot | V_{1:m}, X_{1:n})
\end{equation}

\section{SPD for MLLMs}
\begin{figure}
    \centering
    \includegraphics[width=1\linewidth]{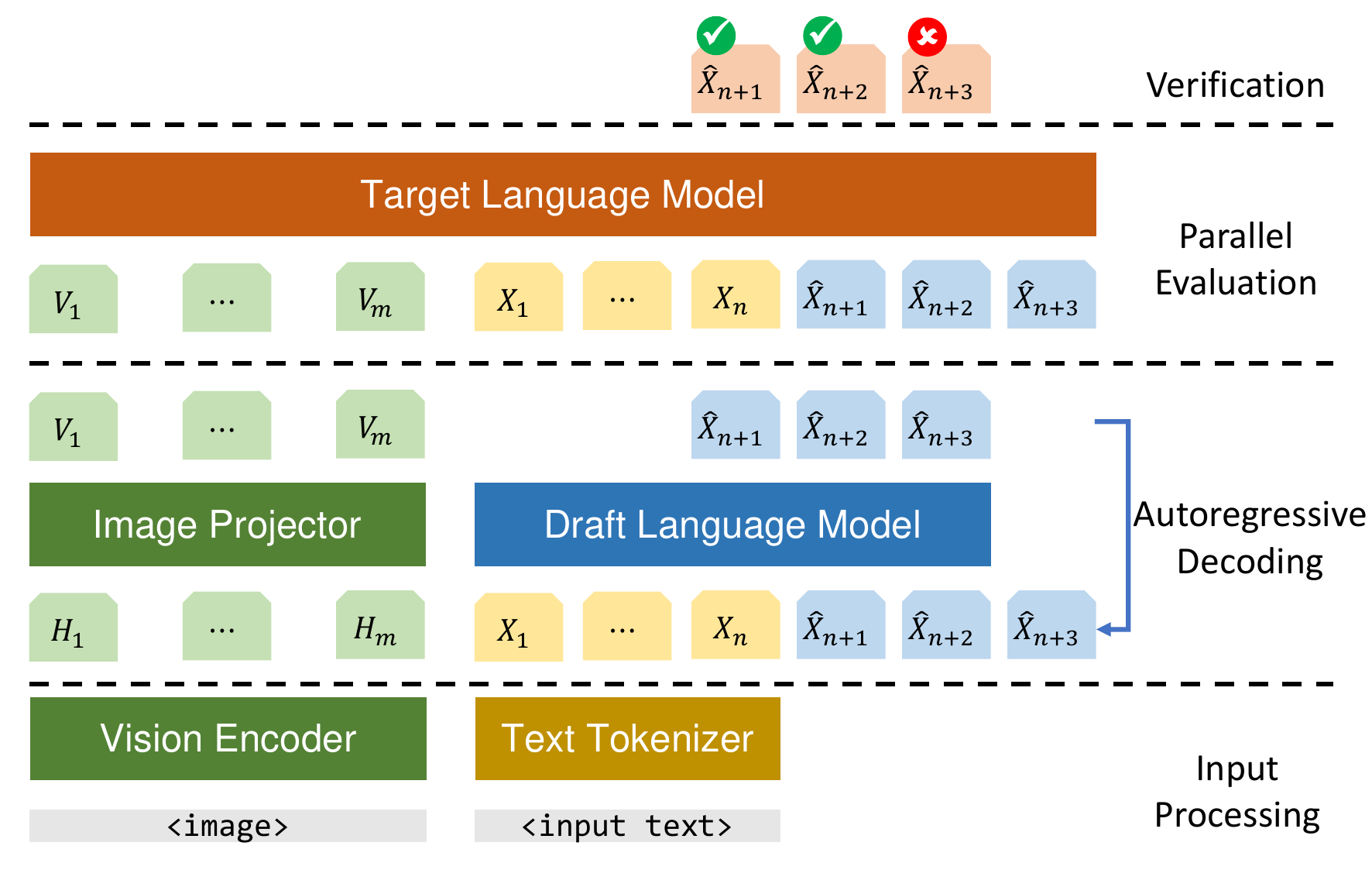}
    \caption{SPD with a MLLM as target having three components: vision encoder, image projector, and target language model, and the smaller language model as draft. The small draft model generates draft tokens autoregressively for block-size number of iterations followed by parallel evaluation by the target language model which also uses image features.}
    \vspace{-0.3cm}
    \label{fig:SPD_block_diagram}
\end{figure}
To achieve higher gain with speculative decoding, we need a draft model significantly smaller than  and well-aligned with our target model (LLaVA-7B). The most common choice for draft models in prior works on LLMs is to use a small pre-trained model from the same family of models as the target model or train a smaller model which has the same architecture as the target model \cite{miao2023specinfer}. Since there is no publicly available smaller model in the LLaVA family, we need to train a draft model from scratch. A natural choice for draft model architecture is to follow LLaVA's architecture where the draft model comprises an adapter and a language-model backbone with smaller number of parameters than the LLaVA 7B. In our approach, we use both, \textit{1) a smaller LLaVA draft model} which consists of a smaller image adapter and a draft language model, and \textit{2) the language-only draft model} which generates draft tokens by conditioning only on the input text tokens. Given an input image with image embeddings $V_{1:m}$, token embeddings $X_{1:n}$ the draft model generates the draft tokens $\hat{X}_{n+1:n+L}$ where the draft token
\begin{align*}
\hat{X}_{n+j} \sim p (\cdot | X_{1:n}, \hat{X}_{n+1:n+j-1})    
\end{align*}
is generated by conditioning only on the text tokens. The target LLaVA model verifies the draft tokens by computing the target distribution which is conditioned on both the image embeddings $V_{1:m}$ and the text token embeddings $X_{1:n}$, i.e., draft token $\hat{X}_{n+j}$ is accepted with probability 
\begin{align*}
    \min\left\{1, \frac{q(\hat{X}_{n+j}|V_{1:m}, X_{1:n}, \hat{X}_{n+1:n+j-1})}{p(\hat{X}_{n+j}|X_{1:n}, \hat{X}_{n+1:n+j-1})}\right\}.
\end{align*}
Using the language-model-only draft model is more efficient than a draft model with LLaVA architecture since \textit{1) it does not need an additional adapter} as it does not condition on the image embeddings for generating draft tokens, and \textit{2) it does not require the training of the adapter}. Figure \ref{fig:SPD_block_diagram} shows SPD with MLLM consisting of the smaller draft language model doing  autoregressive generation followed by the large target model evaluating the draft model predicted tokens in parallel while using the image. 

\section{Experiments}
We run experiments on three visual instruction tasks using SPD with LLaVA-7B \cite{liu2023improved} as our target model which uses the LLaMA-7B model as the language-model backbone. We employ draft models that underwent different stages of training with the size of the language part of each draft model fixed to 115M. %The model configuration are present in (cite appendix subsection). 

\paragraph{Draft Model Candidates.}
We train draft model of size $115M$ which follow the LLaMA-2 architecture. We follow the training pipeline of \cite{goel2024direct} to pre-train a draft model from scratch and fine-tune the draft model on instruction finetuning datasets using TVD++ loss \cite{goel2024direct}. We further fine-tune our draft model on a subset of LLaVA Instruct 150K dataset \cite{liu2024visual}. For our experiments, we consider the following four draft models after each stage of training and finetuning: \textit{1) base-LLaMA}, a draft LLaMA model after pre-training using next-token-prediction loss on 600B English tokens, \textit{2) chat-LLaMA}, an instruction fine-tuned draft LLaMA model following \cite{goel2024direct} initialized with base-LLaMA draft model, and \textit{3) fine-tuned-LLaVA} (ft-llava), a fine-tuned LLaVA draft model where the image adapter was initialized using subcloning \cite{samragh2023weight} of LLaVA-7B image adapter and the language model was initialized from the chat-LLaMA draft model (the model is then fine-tuned on LLaVA dataset). We also include another draft model \textit{4) fine-tuned-LLaVA-text} (ft-llava-text), which simply uses the language model part of \textit{3)}. 
Note that only the fine-tuned-LLaVA draft model uses image information while all other draft models only consume the text part of the input prompt; when the draft model uses image information, the vision encoder (CLIP-based \cite{radford2021learning}) is shared with the target model to avoid re-computation of image embeddings. The detailed parameters are given in Appendix \ref{app:model_config}

\paragraph{Evaluation Tasks.}
We focus on open-ended text generation and multiple choice question-answering with reasoning to encourage a higher number of token generation, which is beneficial when using SPD. To this end, we evaluate on 1) \textbf{LLaVA Instruct 150K dataset} \cite{liu2024visual}, 2) Image captioning task on images from \textbf{COCO dataset} \cite{lin2014microsoft}, and 3) \textbf{Science QA (SQA)} with chain-of-thought (CoT) reasoning \cite{lu2022learn}. The system prompts settings for all the tasks  are described in Appendix \ref{app:sys_prompts}

\paragraph{Metrics.}
The efficacy of SPD is evaluated with the following metrics; 1) \textbf{block efficiency} ($\tau$), the average number of tokens generated per block (or target model run), for a block of size $\gamma$ and input $x$, the maximum value of $\tau(x)$ can be $\gamma + 1$, block-size ($\gamma$) is also known as draft length (DL) in some works; 2) memory-bound speedup (\textbf{MBSU}), the hypothetical speedup achieved by SPD for a given block efficiency $\tau(x)$ and a relative latency $c$ defined as ratio between number of parameters of draft to target model, i.e., $\mathrm{MBSU}(x)=\frac{c\tau(x)}{c\gamma + 1}$; 3) \textbf{token rate}, the total number of tokens generated divided by the total time of generation, giving an estimate of tokens generated for per second. We measure these metrics on various tasks using different block size $\gamma$ in $\{3, 5\}$

\begin{figure}
    \centering
    \includegraphics[width=\linewidth]{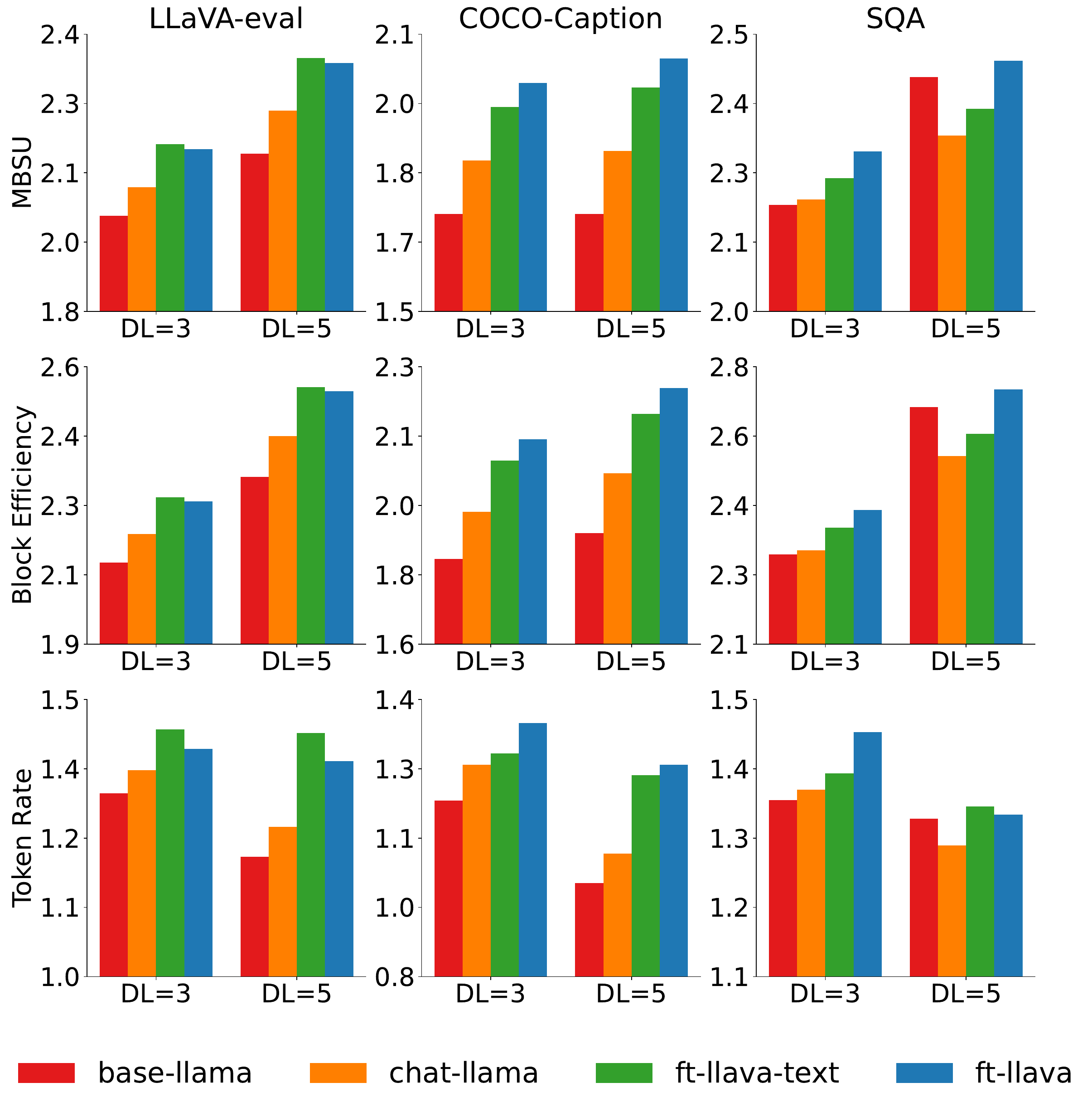}
    \caption{MBSU, block efficiency and token rate (relative to auto-regressive generation) for SPD are depicted; We consider LLaVA-eval, COCO-Caption and SQA datasets for evaluation; For draft models, base-LLaMA, chat-LLaMA, ft-LLaVA-text, ft-LLaVA are considered, we consider three text-only draft models and a single text and image draft model; For draft length (DL) (or block size of SPD), we consider either 3 or 5.}
    \label{fig:result}
    \vspace{-0.1in}
\end{figure}

\paragraph{Decoding.} We use greedy decoding for all experiments so that the SPD generation is identical to the target model's autoregressive generation. We leave it as future work to explore sampling-based decoding (varying temperature, varying top-$p$, top-$k$) in the context of SPD for MLLMs. 

\begin{figure}[t]
\begin{tcolorbox}[colback=blue!1!white,colframe=blue!50!black]
\footnotesize
\begin{wrapfigure}{L}{0.5\linewidth}
    \vspace{-0.5cm}
  \includegraphics[width=1\linewidth]{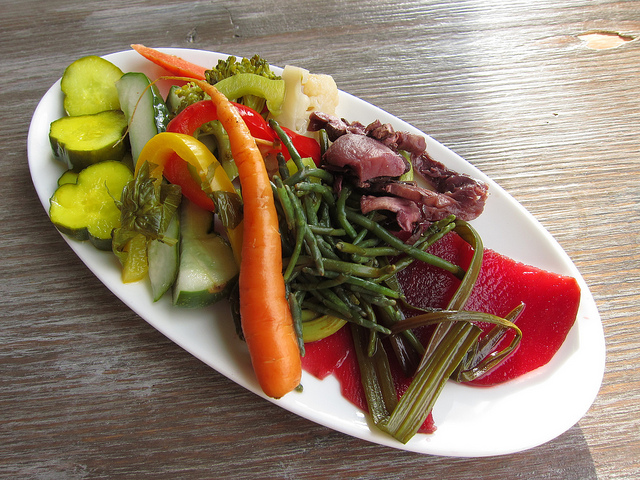}
  \caption{Example 1}
  \label{fig:COCO_1}
  \vspace{-0.4cm}
\end{wrapfigure}
\textbf{Input prompt}:
\textlangle s\textrangle A chat between a curious user and an artificial intelligence assistant. The assistant gives helpful, detailed, and polite answers to the user's questions.  USER: \textlangle image\textrangle\\ 
Provide a detailed description of the given image  ASSISTANT:
\\
\textbf{Output generation}:
The \Blue{image} features \Blue{a} white plate filled \Blue{with a} variety \Blue{of} vege\Blue{tables, including} several car\Blue{rots,} green \Blue{beans,} and bro\Blue{ccoli.} The car\Blue{rots are} scattered across \Blue{the plate,} with some placed closer \Blue{to the} center and \Blue{others} towards \Blue{the} edges. \Blue{The green} be\Blue{ans are} also spread \Blue{out,} with \Blue{some} near \Blue{the center} and \Blue{others} closer \Blue{to the edges}. \Blue{The brocc}oli \Blue{is} located towards \Blue{the} top left corner \Blue{of the plate}. \Blue{The} arrangement \Blue{of the veget}ables creates \Blue{a} color\Blue{ful and} app\Blue{etizin}g display.
    % \begin{minipage}[t]{0.5\linewidth}
    % \vspace*{0pt}
    % yoyoyoyo
    % \end{minipage}\hfill%
    % \begin{minipage}[t]{0.4\linewidth}
    % \vspace*{0pt}
    %     \includegraphics[height=0.6\textheight,width=\linewidth]{images/coco/coco_294.png}
    %     \captionof{figure}{Misura della carica elettrica}\label{fig:fig1}
    % \end{minipage}
\end{tcolorbox}
\begin{tcolorbox}[colback=blue!1!white,colframe=blue!50!black]
\footnotesize
\begin{wrapfigure}{L}{0.55\linewidth}
    \vspace{-0.5cm}
  \includegraphics[width=1\linewidth]{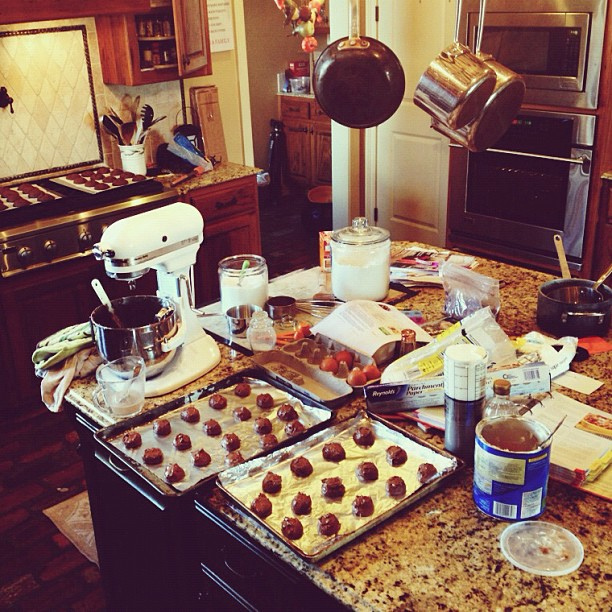}
  \caption{Example 2}
  \label{fig:COCO_2}  
  \vspace{-0.4cm}
\end{wrapfigure}
\textbf{Input prompt}:
\textlangle s\textrangle A chat between a curious user and an artificial intelligence assistant. The assistant gives helpful, detailed, and polite answers to the user's questions.  USER: \textlangle image\textrangle\\ 
Provide a detailed description of the given image  ASSISTANT:
\\
\textbf{Output generation}:
\Blue{The image} dep\Blue{icts a} kitchen counter filled \Blue{with various food} items and kitchen \Blue{utensils}. There \Blue{are several} bow\Blue{ls} placed \Blue{on the counter}, with one large \Blue{bowl} towards \Blue{the} left side and two \Blue{smaller bowls} closer \Blue{to the} center. \Blue{A} few cu\Blue{ps} can \Blue{be} seen as well, with one near \Blue{the} left side \Blue{of the counter} and another towards \Blue{the right} side. 

In addition \Blue{to the bowl}s and cu\Blue{ps, there} are multiple bott\Blue{les} scattered across \Blue{the counter}, with some placed near \Blue{the} center and \Blue{others} towards \Blue{the right} side. A sp\Blue{oon is} also visible \Blue{on the counter}, located \Blue{near the} center. 

\Blue{The} kitchen \Blue{counter is} surrounded \Blue{by} various app\Blue{liances}, including an \Blue{oven} on \Blue{the} right side, \Blue{a} sink in \Blue{the} background, \Blue{and a} re\Blue{frigerator} on \Blue{the left} side. A mic\Blue{rowave} can \Blue{be seen} above \Blue{the counter}, and \Blue{a} kn\Blue{ife is placed} near the right edge \Blue{of the counter}.

\end{tcolorbox}
\caption{SPD examples on COCO-caption task}
\vspace{-0.2in}
\label{fig:qualitative_example}
\end{figure}

\paragraph{Results.}

Our results show that using SPD with LLaVA 7B target model gives considerable speedup in output generation, and we emphasize that when using a draft model without any image information, SPD can still give considerable and competitive speedup to that of a draft model using image information. 

From Figure \ref{fig:result} (top and middle plots), we observe that using SPD gives more than $2\times$ gains in terms of block efficiency and MBSU. The performance trend when increasing the block size from 3 to 5 for each task is similar with the exception for SQA task where base-llama draft model performs better than other text-only draft models for block size $=5$. For LLaVA-eval task on both block sizes (either 3 or 5), the ft-llava-text draft model performs the best closely followed by ft-llava. For COCO-caption task, ft-llava performs best, followed by ft-llava-text for both block sizes. Lastly, for the SQA task, for block size=3, ft-llava draft model performs the best followed by ft-llava-text while for block-size=5, ft-llava draft model performs best followed by base-llama.
In addition, all our draft models show the improved token rate compared to auto-regressive generation in Figure \ref{fig:result} (bottom) with block size 3 giving better token rate than block size 5, thus, SPD generates more token tokens per second than autoregressive decoding. The token rate shown corresponds to the ratio of the token rate of SPD using a particular draft model to the token rate of autoregressive generation using target model.  

We further provide qualitative results on the COCO-captioning task to show the tokens accepted during a generation process when using the fine-tune-LLaVA-text draft model so no image information used by draft model in Figure \ref{fig:qualitative_example}. Based on the output generations in the figure, where tokens in blue and underlined are the accepted tokens, we observe that the draft model can predict common words and propositions, along with halves of words. For example, the draft model can predict ``tables" given ``vege".
Similarly in the second example, given the context and additional token ``app", the draft model was able to predict ``liances". We believe in general open-ended text generation has several tokens comprising of common words, propositions, and word completions which do not require knowledge of image tokens, thus, even a draft model without using image information gives competitive performance. Moreover, draft model can also predict the repetition of certain tokens once they have been generated. For example, in the second image the word ``counter" and ``bowls" can be predicted by the draft model multiple times once it has been generated by the target model. Lastly, performing more rigorous training on a small multi-modal language model is left as our future work.

\section{Conclusion}
In this paper, we present the first effort towards using speculative decoding for accelerating inference when using multi-modal large language models, specifically for image-text domain. We show that using the text-only draft model achieves performance competitive to using a draft model utilizing image features. We perform various experiments on different visual question-answering tasks focusing on generating higher number output tokens: open-ended text generation and text generation with reasoning using different draft models (text-only and image-text). We achieved significant speedup of upto $2.37\times$ for text-only draft model and marginal better speedup for image-text draft model, empirically showing the potential of using SPD for MLLMs.  

Our work opens several future avenues owing to the general framework presented. Our work can be extended to other target models such as BLIP-2 \cite{li2023blip}, MiniGPT-4 \cite{zhu2023minigpt} and OpenFlamingo \cite{awadalla2023openflamingo}, and other modalities such as audio \cite{chu2023qwen} which are also bottlenecked by auto-regressive generation.  Furthermore, recent advancement in SPD algorithm to tree-based decoding can also be used following~\cite{sun2023spectr,miao2023specinfer,medusa,jeon2024recursive} to further increase generation speed.

{
    \small
    \bibliographystyle{ieeenat_fullname}
    \bibliography{main}
}

\appendix
\onecolumn
\section{Appendix}
\subsection{Model Configurations}
\label{app:model_config}
The LLaVA-7B model uses: (i) vision encoder, (ii) multi-layer perceptron (MLP) based image adapter/projector, and (iii) LLaMA 7B language model. 
The visual encoder is CLIP ViT-L/14 with details present in  \cite{radford2021learning}, the MLP-based image adapter has $2$ linear layer with following sizes: $1024\times4096$ and $4096\times 4096$. For the scenario when draft model also has image adapter the sizes are $1024 \times 1024$ and $1024 \times 1024$.

The following configurations are used for our target and draft language model part which follows the LLaMA architecture:
\begin{table}[h]
    \centering
    \caption{Draft and target model configurations}
    \begin{tabular}{l||r|r}
    
           & target (7B) & draft (115M) \\
           \hline
    Layers & 32 & 4\\
    Attention heads & 32 & 8\\
    Intermediate dim & 11,008 & 2,816\\
    Hidden dim & 2,048 & 1,024\\
    Activation & SiLU & SiLU
    \end{tabular}
    \label{tab:model_config}
\end{table}

\subsection{System Prompts}
\label{app:sys_prompts}
We use the following systems prompts for the respective task. The special image token is used to include the image data (\textit{$<$image$>$}) 

\textbf{LLaVA-eval.} We follow the prompt style given in \cite{liu2024visual}, LLaVA has multiple questions and responses which we divide into different samples. 

\textit{$<$s$>$ A chat between a curious user and an artificial intelligence assistant. The assistant gives helpful, detailed, and polite answers to the user's questions.  USER: $<$image$>$ $\\$Question $Q_{1}$  ASSISTANT: response $R_{1}$. USER: Question $Q_{2}$ $\dots$}.

\textbf{COCO-caption.} As COCO dataset doesn't have any question prompts, we prompted the model with a prompt similar to above.

\textit{$<$s$>$ A chat between a curious user and an artificial intelligence assistant. The assistant gives helpful, detailed, and polite answers to the user's questions.  USER: $<$image$>$ $\\$Provide a detailed description of the given image  ASSISTANT:}

\textbf{Science QA.} We follow the prompt style provided in \cite{lu2022learn} with a single in-context example of the question, choices, answer and reasoning to enable Chain-of-Thought (CoT) reasoning. Additionally we only consider the test samples which have an associated image. 

\begin{align*}
    & \text{Question: question :} I_{i}^{ques} \\
    & \text{Options: (0) option : } I_{i1}^{opt} \text{ (1) option : } I_{i2}^{opt} \text{ (2) option : } I_{i3}^{opt} \\
    & \text{Context: context : } I_{i}^{cont} \\
    & \text{Answer: The answer is } I_{i}^{ans} \text{. BECAUSE: lecture} I_{i}^{lect} \text{ explanation : } I_{i}^{exp} \\
    \\
    & <image>
    \\
    & \text{Question: question :} I_{test}^{ques} \\
    & \text{Options: (0) option : } I_{test,1}^{opt} \text{ (1) option : } I_{test,2}^{opt} \text{ (2) option : } I_{test,3}^{opt} \\
    & \text{Context: context : } I_{test}^{cont} \\
    & \text{Answer: The answer is}     
\end{align*}

% \begin{align*}
%     & \text{Question: } I^{ques} \\
%     & \text{Options: (0) } I_{1}^{opt} \text{ (1) } I_{2}^{opt} \text{ (2) } I_{3}^{opt} \\
%     & \text{Context: } I^{cont} \\
%     & \text{Answer: The answer is } I^{ans} \text{. BECAUSE: lecture}
% \end{align*}
where, the subscript $i$ is for in-context example. 

In the SQA paper, the context field is provided by generating a caption for the associated image using an image captioning model, however, these captions were often simple and didn't provide a detailed description of the image which is needed for answering the question. For this reason, the context field is  filled with ``hint" field provided in the SQA dataset. For the in-context sample we choose a sample without any associated image as the target LLaVA 7B cannot consume multiple images. We leave it as a future work to experiment SPD with more than $1$ in-context examples.

\end{document}